\documentclass[conference]{IEEEtran}
\IEEEoverridecommandlockouts
\usepackage{cite}
\usepackage{amsmath,amssymb,amsfonts}
\usepackage{booktabs} 
\usepackage{algorithmic}
\usepackage{graphicx}
\usepackage{textcomp}
\usepackage{xcolor}
\usepackage{multicol}
\usepackage{multirow}
\usepackage{url}
\usepackage{subcaption}  
\usepackage{changes}
\def\BibTeX{{\rm B\kern-.05em{\sc i\kern-.025em b}\kern-.08em
    T\kern-.1667em\lower.7ex\hbox{E}\kern-.125emX}}
\begin{document}

\title{Laplacian Frequency Interaction Network for Rural Thematic Road Extraction\\

\thanks{This research was financially supported by the National Key R\&D Program of China (Grant No. 2025YFE0103600), the National Natural Science Foundation of China (Grant No. 32301691), and the 2115 Talent Development Program of China Agricultural University.
*Corresponding author.}
}

 \author{\IEEEauthorblockN{1\textsuperscript{st} Baiyan Chen}
 \IEEEauthorblockA{\textit{College of Information and Electrical Engineering} \\
 \textit{China Agricultural University}\\
 Beijing, China \\
 baiyanchen@cau.edu.cn}
 \and
 \IEEEauthorblockN{2\textsuperscript{nd} Weixin Zhai\textsuperscript{*}}
 \IEEEauthorblockA{\textit{College of Information and Electrical Engineering} \\
 \textit{China Agricultural University}\\
 Beijing, China \\
 zhaiweixin@cau.edu.cn}
 }

\maketitle

\begin{abstract}
Rural thematic road network construction aims to extract topological road structures from movement trajectory images of agricultural machinery. However, this task faces challenges where downsampling methods commonly used in existing studies tend to blur the sparse high-frequency road structures, and the heavy noise from dense field operations often leads to fragmented or redundant topologies in the extracted networks. To address these challenges, we propose LFINet, a Laplacian Frequency Interaction Network. The network begins with a Laplacian Multi-scale Separator (LMS) to decouple the image into low-frequency semantic contexts and high-frequency structural details. These components are then processed by the Cross-Frequency Interaction Block (CFIB) through a dual-pathway architecture in which a High-Frequency Block (HFB) refines local structures while a Spatial Transformer (ST) captures global semantics. Subsequently, a Frequency Gated Modulation (FGM) mechanism integrates the features from pathways by leveraging semantic contexts to calibrate the structural details. Finally, a Progressive Reconstruction Decoder iteratively fuses multi-scale features to ensure topological consistency. Experiments conducted on a real-world agricultural trajectories dataset from Henan Province, China, show that LFINet establishes a new state-of-the-art. Specifically, it achieves an F1-score of 92.54\% and an IoU of 86.12\%, surpassing the second-ranked method by 0.64\% and 1.13\%, respectively. This confirms its capability to effectively construct topological road networks from noisy and sparse field data.

\end{abstract}


\section{Introduction}
Rural thematic road networks are essential for precision agriculture (e.g., agricultural machinery navigation and route planning) but are poorly covered by general GPS maps. Traditional remote sensing also fails due to the irregular morphologies of rural roads and their textural similarities with fields. In contrast, agricultural machinery trajectory data offer a superior alternative. The widespread deployment of GNSS-equipped agricultural machinery has generated vast archives of movement data, recording not only field operations but also the transit paths connecting farmlands. To leverage these data for road extraction, discrete trajectory points are typically rasterized into trajectory-derived images, facilitating pixel-level delineation of valid road segments.
However, processing these images is challenging due to the inherent complexity of agricultural machinery movement, which includes irregular field operations, navigation through narrow roads, transit across wide open fields. Such behavioral complexity complicates the distinction between valid roads and field paths.

The first challenge arises from the structural discontinuity caused by sparse sampling. Discrete GNSS logging intervals inherently result in sparse spatial distributions, creating significant gaps between trajectory points, particularly in high-speed straight sections. Existing convolutional neural network (CNN)-based methods (e.g., U-Net \cite{b_unet}, DeepLabV3+ \cite{b_deeplabv3plus}) rely on local convolutional operators for feature extraction. More advanced architectures, such as Swin-UNet \cite{b_swin_unet} and SegFormer \cite{b_segformer}, integrate Transformer mechanisms for enhanced global context modeling. However, these local operations are ill-suited for sparse trajectories, as the inter-point distances often exceed the network's effective receptive field \cite{b1}. Similarly, Vision Mamba-based methods \cite{b18_pathmamba} leverage state space models to capture global context with linear complexity, yet their sequential scanning mechanisms remain fundamentally misaligned with the irregular, non-sequential nature of sparse trajectory distributions.

The second challenge involves the absence of feature domain differentiation. In trajectory images, high-frequency components represent thin linear road structures, while low-frequency components encode global connectivity. Current frequency-aware models \cite{b20_transimgs} and dual-decoder architectures \cite{b_DGMAp, b15} still process these distinct signals simultaneously in a common mixed-frequency encoder. In these architectures, both the broad semantic background (low-frequency) and the sharp, thin road boundaries (high-frequency) are processed simultaneously. Consequently, critical high-frequency structural information is severely attenuated during downsampling operations. Consequently, critical high-frequency structural details are severely attenuated during downsampling, restricting the model's ability to preserve thin edges while capturing global topology.

\begin{figure*}[t]
    \centering
    \includegraphics[width=\textwidth]{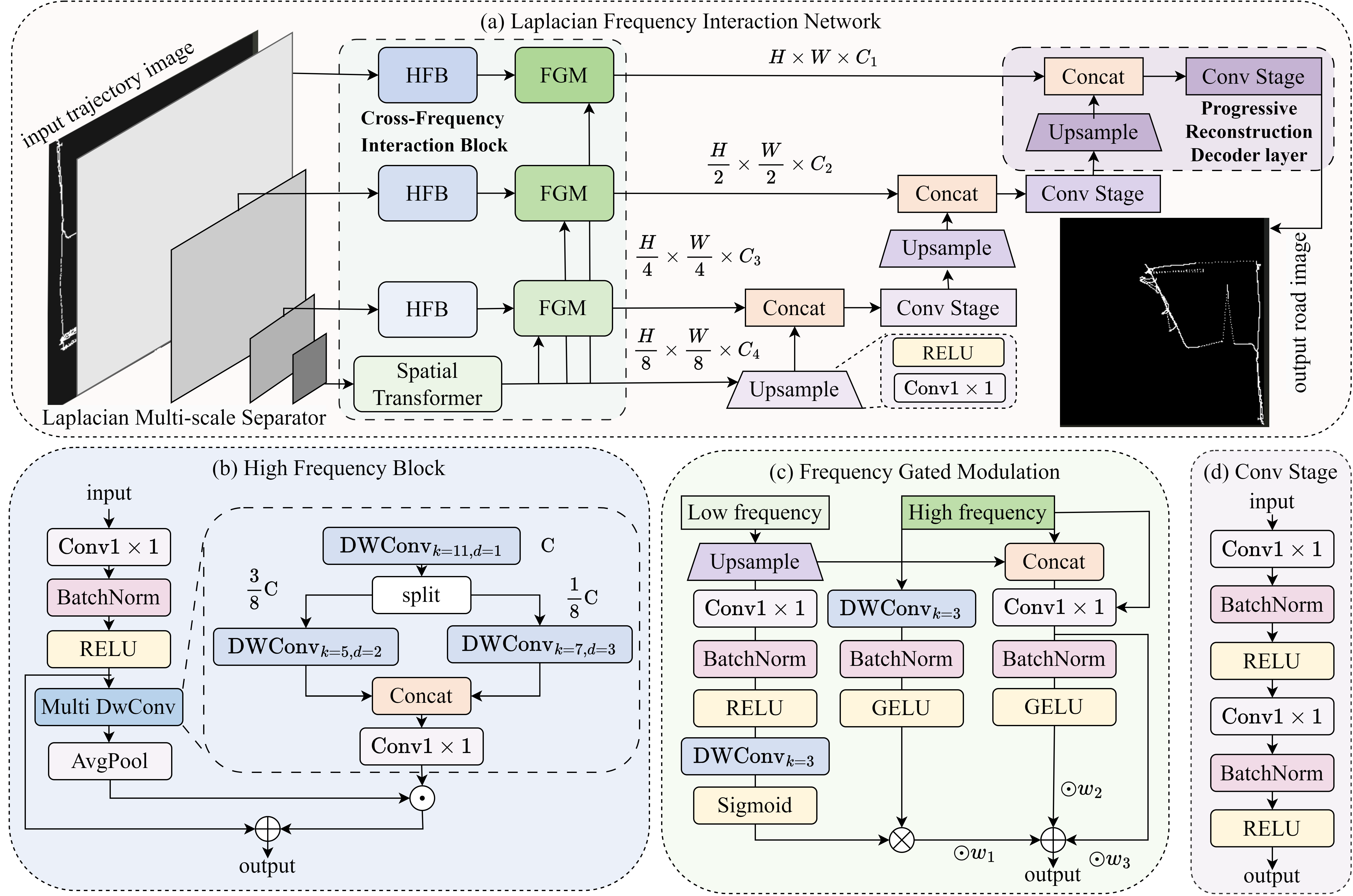}
    \caption{Overview of the LFINet. (a) The main pipeline consists of the Laplacian Multi-scale Separator, Cross-Frequency Interaction Block (CFIB), and Progressive Reconstruction Decoder (PRD). (b) The High-Frequency Block (HFB) utilizes cascaded depth-wise convolutions for multi-scale feature extraction. (c) The Frequency Gated Modulation (FGM) module synergizes low- and high-frequency components using semantic gating. (d) Detailed structure of the Conv Stage.}
    \label{fig:overview}
\end{figure*}

The third challenge stems from dual-source noise interference. Trajectory data suffer from dual-source interference, which includes both random perturbations such as GNSS positioning jitter and structural noise arising from off-center field operations and dense overlapping points at intersections. Hand-crafted post-processing (e.g., morphological operations \cite{b22_mor}, CRF \cite{b23_CRF}) often over-smooths genuine road boundaries. Conversely, while modern deep learning models \cite{b10} eliminate manual intervention, they lack explicit semantic priors to differentiate between true road structures and task-irrelevant agricultural interference, struggling to filter out structural noise.

To address these challenges, we propose LFINet, a Laplacian Frequency Interaction Network for road extraction. LFINet first explicitly decouples input images into low-frequency semantic skeletons and multi-scale high-frequency structural details via a Laplacian Multi-scale Separator (LMS). To bridge the spectral gap, a Cross-Frequency Interaction Block (CFIB) is designed to synergize global connectivity and local edge details. Finally, a Progressive Reconstruction Decoder (PRD) iteratively integrates these calibrated features to ensure topological consistency in the extracted road networks.

Our contributions are three-fold:
\begin{itemize}
\item We propose a Laplacian Multi-scale Separator (LMS) that explicitly decouples frequency-specific representations, preventing the attenuation of high-frequency structural components during feature encoding while ensuring the integrity of fine-grained road boundaries.
\item We design a Cross-Frequency Interaction Block (CFIB) to synergize local geometric perception and global semantics. It integrates cascaded dilated convolutions and a Spatial Transformer (ST) via a Frequency Gated Modulation (FGM) mechanism, adaptively suppressing background noise using semantic priors.
\item We propose a Progressive Reconstruction Decoder (PRD) that iteratively aggregates refined features, ensuring topological continuity and effectively mitigating the spatial detail loss typical of encoder-decoder architectures.
\end{itemize}

\section{Methods}

\subsection{Overall Framework}

The proposed LFINet adopts a ``decompose-interact-reconstruct'' strategy. As illustrated in \figurename~\ref{fig:overview}, the framework consists of three core stages. In the frequency decomposition stage, the Laplacian Multi-scale Separator (LMS) factorizes the input trajectory image into a Gaussian-based low-frequency semantic base ($I_{\mathrm{lf}}$) and a hierarchy of Laplacian high-frequency detail components ($L_0, L_1, L_2$). During the cross-frequency interaction stage, the Cross-Frequency Interaction Block (CFIB) processes these components through two parallel pathways, including a High-Frequency Block (HFB) that utilizes cascaded dilated convolutions to refine local edge perception and a Spatial Transformer (ST) that captures long-range global dependencies within the low-frequency semantics. These pathways are then synergistically modulated by the Frequency Gated Modulation (FGM) module, which uses semantic priors to calibrate high-frequency details. Finally, the Progressive Reconstruction Decoder (PRD) restores spatial resolution through a hierarchical upsampling pathway. By iteratively fusing refined cross-frequency features via skip connections, the PRD ensures that the final output maintains high topological integrity and spatial fidelity.

\subsection{Laplacian Multi-scale Separator}
To prevent the blurring of fine-grained trajectory boundaries caused by the implicit frequency mixing in CNN-based backbones, we propose a Laplacian Multi-scale Separator (LMS), which decomposes the input trajectory image $I\in \mathbb{R}^{H\times W\times C}$ into multi-scale frequency components for subsequent refinement.
The LMS adopts a recursive ``blur-and-subtract'' decomposition strategy. First, a Gaussian pyramid $\{I_l\}$ represents the image at varying scales, capturing semantic information at progressive scales.
\begin{equation}
I_{l+1} = \text{Pool}_{2 \times 2}(I_l * K_g), \quad l = 0, \dots, 3,
\end{equation}
where $I_0$ is the input image, $K_g$ denotes a fixed $3 \times 3$ Gaussian kernel for anti-aliasing, and $\text{Pool}_{2 \times 2}$ represents average pooling for downsampling.
Subsequently, to isolate specific frequency bands and preserve edge details, we extract each high-frequency Laplacian level $L_l$ by subtracting the upsampled version of the coarser level from the current resolution features.
\begin{equation}
L_l = I_l - \text{Interp}_{2 \times 2}(I_{l+1}), \quad l = 0, 1, 2,
\end{equation}
where $\text{Interp}_{2 \times 2}$ signifies bilinear interpolation upsampling.
As a result, the input is transformed into a set of multi-scale high-frequency features $\{L_l\}_{l=0}^2$, where $L_0 \in \mathbb{R}^{H \times W \times C}$, $L_1 \in \mathbb{R}^{\frac{H}{2} \times \frac{W}{2} \times C}$, $L_2 \in \mathbb{R}^{\frac{H}{4} \times \frac{W}{4} \times C}$, and a low-frequency feature $I_{\mathrm{lf}} \in \mathbb{R}^{\frac{H}{8} \times \frac{W}{8} \times C}$ that captures global semantics. Specifically, $L_0$ preserves fine-grained textures and local topological details, while $L_1$ and $L_2$ capture intermediate structural features at corresponding scales. This frequency-aware separation allows the subsequent modules to utilize $I_\mathrm{lf}$ for maintaining the connectivity of fragmented paths, while leveraging $L_l$ to delineate road boundaries against complex field backgrounds.

\subsection{Cross-Frequency Interaction Block (CFIB)}
Low-frequency features capture overall road connectivity, while high-frequency features delineate fine boundary details.
To effectively utilize the decomposed frequency components, the Cross-Frequency Interaction Block (CFIB) with a dual-pathway architecture is developed to process low-frequency semantics and high-frequency details independently.
Specifically, the low-frequency base $I_\mathrm{lf}$ is fed into a Spatial Transformer, which utilizes a lightweight Transformer to capture long-range global dependencies and semantic contexts. Simultaneously, the high-frequency components $L_l$ are processed by the High-Frequency Block (HFB) to refine local boundary details. Finally, the outputs from both pathways are synergistically integrated via the Frequency Gated Modulation (FGM) module.
\subsubsection{Spatial Transformer (ST)}
This module leverages self-attention mechanisms to establish direct interactions between spatially distant features. By processing the low-frequency representation as a sequence of feature tokens, the Transformer dynamically aggregates semantic contexts across the global spatial domain. This capability enables the network to perceive overall connectivity patterns, effectively suppressing local ambiguities (e.g., field textures resembling roads) by verifying their consistency with the global road topology. Consequently, the generated output serves as a robust semantic prior, guiding the subsequent FGM module to distinguish genuine road structures from background noise. 
\subsubsection{High-Frequency Block (HFB)}
For one pathway, the HFB ensembles cascaded depthwise convolutions (DWConv) to expand the receptive field while preserving high-resolution spatial information.
First, a convolution is applied to the input Laplacian feature $L_l$ to map it into a high-dimensional feature space, enhancing its representational capacity.
\begin{equation}
\tilde{L}_l = \text{ReLU}(\text{BN}(\text{Conv}_{3 \times 3}(L_l))).
\end{equation}

Then, $\tilde{L}_l$ undergoes multi-order DWConv, where three parallel DWConv layers with dilation ratios $d \in \{1, 2, 3\}$ are used to capture multi-scale interactions.
A large-kernel depthwise convolution ($11 \times 11$) is first performed on the global features for base spatial feature extraction.
\begin{equation}
{F}_0 = \text{DWConv}_{k=11, d=1}(\tilde{L}_{l}).
\end{equation}

Subsequently, ${F}_0$ is sliced into three sub-feature sets $[{F}_a, {F}_b, {F}_c]$ along the channel dimension with a ratio of 1:3:4.

These subsets are processed hierarchically to encode structural patterns at varying scales. ${F}_a$ retains the base information. The remaining subsets are refined using depthwise convolutions with increasing kernel sizes and dilation rates, effectively addressing the continuity issues of slender road structures.
\begin{equation}
{F}_1 = \text{DWConv}_{k=5, d=2}({F}_b),
\end{equation}
\begin{equation}
{F}_2 = \text{DWConv}_{k=7, d=3}({F}_c).
\end{equation}

The multi-scale features are then concatenated and aggregated via a pointwise convolution to restore channel interaction.
\begin{equation}
L_\mathrm{cat} = \text{Conv}_{1 \times 1}(\text{Concat}({F}_a, {F}_1, {F}_2)).
\end{equation}

Finally, to suppress background noise and highlight informative feature maps, we employ a lightweight channel-wise reweighting block to modulate the aggregated high-frequency features. The final output $L_l$ is generated by adding the reweighted features to the initial projected feature $\tilde{L}_l$ via a residual connection.
\begin{equation}
L_{l} = \left( L_\mathrm{cat} \odot \sigma(\phi_2(\delta(\phi_1(\mathcal{A}(L_\mathrm{cat}))))) \right) + \tilde{L}_l,
\end{equation}
where $\mathcal{A}$ denotes Global Average Pooling, $\phi_1$ and $\phi_2$ represent the two $1 \times 1$ convolutions (channel reduction and expansion) and $\delta$ is the GELU activation function.

\begin{table}[t]
\caption{Quantitative Comparison of Different Methods}
\label{tab:comparison}
\centering

\setlength{\tabcolsep}{3.5pt} 
\begin{tabular}{lccccccc}
\toprule
\textbf{Methods} & \textbf{F1} & \textbf{IoU} & \textbf{Prec.} & \textbf{Recall} & \textbf{PSNR} & \textbf{Params} & \textbf{FLOPs} \\
& (\%) & (\%) & (\%) & (\%) & (dB) & (M) & (G) \\
\midrule
SAM-Road     & 81.89 & 69.33 & 87.70 & 76.80 & 26.17 & 86.60 & 32.50 \\
DeepMG       & 88.08 & 78.70 & 95.56 & 81.68 & 28.28 & 63.08 & 153.66 \\
T2R-pix2pix  & 91.89 & 84.99 & 97.74 & 86.70 & 30.26 & 42.30 & 18.88\\
AD-HRNet     & 89.04 & 80.25 & 95.68 & 83.26 & 28.17 & 19.41 & 6.26 \\
T2R-GAN      & 87.72 & 78.13 & 94.09 & 82.17 & 27.57 & 6.19 & 12.55\\
T2R2M        & 87.69 & 78.08 & 94.15 & 82.06 & 27.01 & 37.77 & 11.07 \\
NL-LinkNet   & 87.16 & 77.25 & 94.14 & 81.15 & 26.99 & 21.82 & 7.86 \\
Spd-LinkNet  & 86.80 & 76.67 & 94.71 & 80.10 & 27.10 & 39.05 & 8.58 \\
MADSNet      & 75.71 & 60.91 & 71.37 & 80.60 & 22.49 & 28.99 & 8.36 \\
SCSegamba    & 52.37 & 35.47 & 74.02 & 40.51 & 8.66  & 2.8 & 18.16\\
\textbf{LFINet} & \textbf{92.54} & \textbf{86.12} & \textbf{98.60} & \textbf{87.18} & \textbf{31.14} & 22.11  & 65.64 \\
\bottomrule
\end{tabular}
\end{table}

\begin{table}[t]
\caption{Ablation Study of the Proposed LFINet}
\label{tab:ablation}
\centering
\setlength{\tabcolsep}{2.5pt} 

\begin{tabular}{lccccc}
\hline
\textbf{Variants} & \textbf{F1} (\%)& \textbf{IoU} (\%) & \textbf{Prec.} (\%) & \textbf{Recall} (\%) & \textbf{PSNR} (dB)\\
\hline
w/o FGM    & 90.48 & 82.62 & 97.10 & 84.71 & 28.92 \\
w/o HFB    & 91.90 & 85.02 & 98.50 & 86.14 & 30.38 \\
w/o PRD    & 80.18 & 66.92 & 87.78 & 73.80 & 24.96 \\
w/o LMS    & 91.44 & 84.24 & 98.64 & 85.22 & 30.26 \\
w/o ST    & 90.88 & 83.28 & 98.39 & 84.43 & 29.82 \\
\textbf{LFINet}& \textbf{92.54} & \textbf{86.12} & \textbf{98.60} & \textbf{87.18} & \textbf{31.14} \\
\hline
\end{tabular}
\end{table} 

\subsubsection{Frequency Gated Modulation}
Following the refinement of local boundaries in the HFB, the output high-frequency features $L_l$ retain certain task-irrelevant structural noise, such as erratic machinery movements.  To filter this noise, the Frequency Gated Modulation (FGM) module bridges the spectral gap by calibrating high-frequency details using low-frequency semantic guidance.  As illustrated in \figurename~\ref{fig:overview}, the FGM module achieves this synergistic feature rectification through three parallel information streams:

{Semantic Gating Stream}: 
This stream suppresses background noise (e.g., erratic trajectories) by using low-frequency semantics as a filter. We first align the spatial resolution of the low-frequency feature $I_\mathrm{lf}$ to $L_l$ via bilinear upsampling ($\text{Up}$). A spatial attention mask $G$ is then generated to modulate the enhanced high-frequency features $V$.
\begin{equation}
\begin{aligned}
G &= \sigma\!\left(\text{DWConv}_{3 \times 3}\!\left(\text{GELU}\!\left(\text{BN}\!\left(\text{Conv}_{1 \times 1}\!\left(\text{Up}(I_{lf})\right)\right)\right)\right)\right), \\
V &= \text{GELU}\!\left(\text{BN}\!\left(\text{DWConv}_{3 \times 3}\!\left(L_l\right)\right)\right), \\
M_{\text{gate}} &= G \odot V,
\end{aligned}
\label{eq:gate_calculation}
\end{equation}
where $\sigma$ denotes the sigmoid function, and $\odot$ represents element-wise multiplication.

{Frequency Difference Stream}:
To capture the complementary information and structural discrepancies between frequency bands, we employ a concatenated projection.
\begin{equation}
M_{\text{diff}} = \text{GELU}\left(\text{BN}\left(\text{Conv}_{1 \times 1}\left(\text{Concat}(L_l, \text{Up}(I_\mathrm{lf}))\right)\right)\right)
\end{equation}

{Detail Preservation Stream}:
To maintain the fidelity of the original high-frequency edge details and facilitate gradient flow, a residual path is applied.
\begin{equation}
M_{\text{res}} = \text{Conv}_{1 \times 1}(L_l).
\end{equation}

{Adaptive Fusion}:
Finally, to dynamically aggregate these diverse representations, a channel-wise attention mechanism is employed. An aggregation weight vector $\boldsymbol{w} = [w_1, w_2, w_3]$ is predicted from the sum of the three streams via global average pooling and softmax normalization. The final output $L_l$ is then updated.
\begin{equation}
L_l = w_1 \odot M_{\text{gate}} + w_2 \odot M_{\text{diff}} + w_3 \odot M_{\text{res}}.
\end{equation}

\subsection{Progressive Reconstruction Decoder}
To effectively learn the mapping from sparse agricultural machinery trajectories to continuous road networks and optimize the spatial details of the generated topology, we design a Progressive Reconstruction Decoder (PRD).
The decoder consists of progressive reconstruction layers. In each layer, the input encoder features are processed by an Upsample block implemented as a transposed convolution followed by a ReLU activation, which doubles the spatial resolution. To compensate for the information loss caused by downsampling in the encoder, we introduce a skip connection mechanism. The upsampled feature map is concatenated with the refined high-frequency feature $L_l$ from the FGM module. The fused features are then processed by a Conv Stage (CS) to integrate the semantic and structural details. The Conv Stage consists of two stacked $3 \times 3$ convolutional layers, each followed by Batch Normalization (BN) and ReLU activation.
\begin{equation}
\text{CS}(I) = \text{ReLU}(\text{BN}(\text{Conv}_{3 \times 3}(\text{ReLU}(\text{BN}(\text{Conv}_{3 \times 3}(I)))))).
\end{equation}

The progressive update is defined as follows.
\begin{equation}
I_\mathrm{de}^{l} = \text{CS}(\text{Concat}(\text{Upsample}(I_\mathrm{de}^{l+1}), L_l)).
\end{equation}

\begin{figure*}[!t]
    \centering
    \begin{subfigure}{0.158\linewidth}
        \includegraphics[width=\linewidth]{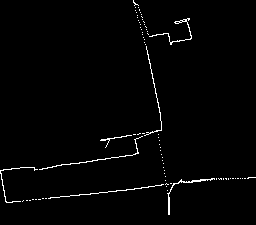}
        \caption{Ground Truth}
    \end{subfigure}
    \hfill
    \begin{subfigure}{0.158\linewidth}
        \includegraphics[width=\linewidth]{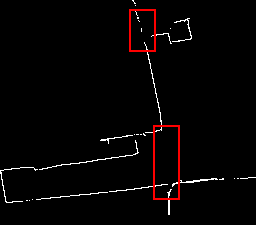}
        \caption{SAM-Road}
    \end{subfigure}
    \hfill
    \begin{subfigure}{0.158\linewidth}
        \includegraphics[width=\linewidth]{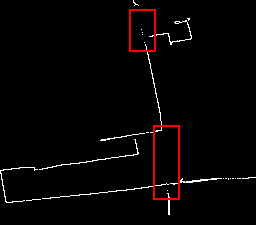}
        \caption{DeepMG}
    \end{subfigure}
    \hfill
    \begin{subfigure}{0.158\linewidth}
        \includegraphics[width=\linewidth]{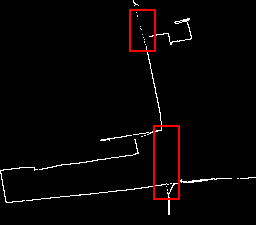}
        \caption{T2R-pix2pix}
    \end{subfigure}
    \hfill
    \begin{subfigure}{0.158\linewidth}
        \includegraphics[width=\linewidth]{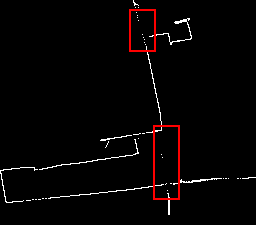}
        \caption{AD-HRNet}
    \end{subfigure}
    \hfill
    \begin{subfigure}{0.158\linewidth}
        \includegraphics[width=\linewidth]{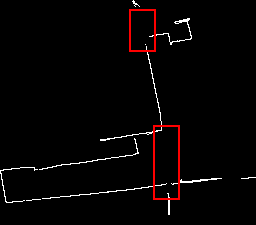}
        \caption{T2R-GAN} 
    \end{subfigure}
    
    \begin{subfigure}{0.158\linewidth}
        \includegraphics[width=\linewidth]{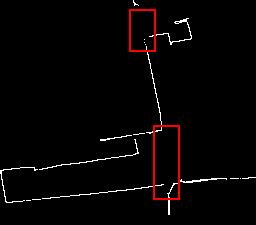}
        \caption{T2R2M}
    \end{subfigure}
    \hfill
    \begin{subfigure}{0.158\linewidth}
        \includegraphics[width=\linewidth]{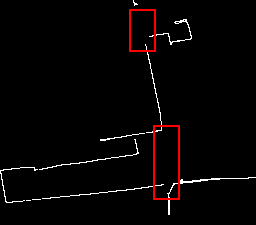}
        \caption{NL-LinkNet}
    \end{subfigure}
    \hfill
    \begin{subfigure}{0.158\linewidth}
        \includegraphics[width=\linewidth]{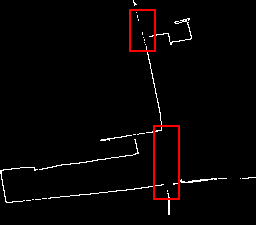}
        \caption{Spd-LinkNet}
    \end{subfigure}
    \hfill
    \begin{subfigure}{0.158\linewidth}
        \includegraphics[width=\linewidth]{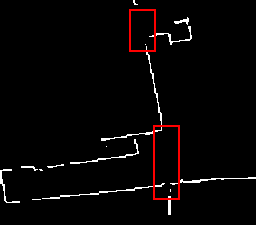}
        \caption{MADSNet}
    \end{subfigure}
    \hfill
    \begin{subfigure}{0.158\linewidth}
        \includegraphics[width=\linewidth]{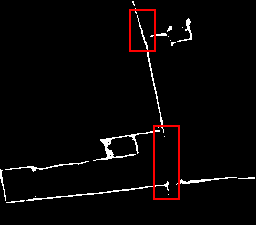}
        \caption{SCSegamba}
    \end{subfigure}
    \hfill
    \begin{subfigure}{0.158\linewidth}
        \includegraphics[width=\linewidth]{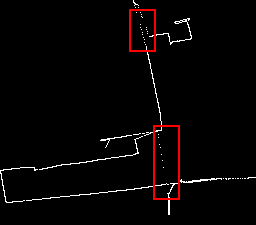}
        \caption{LFINet}
    \end{subfigure}
    
    \caption{Visual comparison with state-of-the-art methods.}
    \label{fig:seven}
\end{figure*}

\begin{figure*}[!t]
    \centering
    \begin{subfigure}{0.137\linewidth}
        \includegraphics[width=\linewidth]{img2/train_1272_gt.png}
        \caption{Ground Truth}
    \end{subfigure}
    \begin{subfigure}{0.137\linewidth}
        \includegraphics[width=\linewidth]{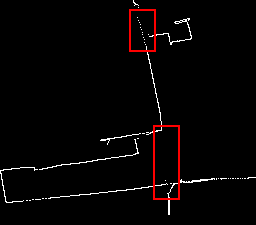}
        \caption{w/o LMS} 
    \end{subfigure}
    \begin{subfigure}{0.137\linewidth}
        \includegraphics[width=\linewidth]{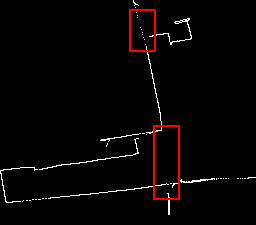}
        \caption{w/o HFB} 
    \end{subfigure}
    \begin{subfigure}{0.137\linewidth}
        \includegraphics[width=\linewidth]{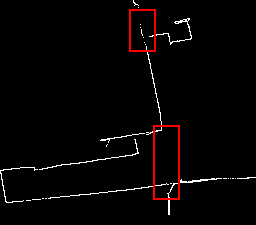}
        \caption{w/o FGM} 
    \end{subfigure}
    \begin{subfigure}{0.137\linewidth}
        \includegraphics[width=\linewidth]{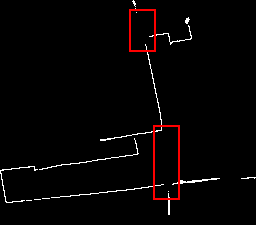}
        \caption{w/o PRD} 
    \end{subfigure}
    \begin{subfigure}{0.137\linewidth}
        \includegraphics[width=\linewidth]{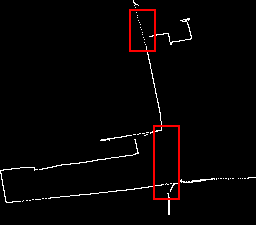}
        \caption{w/o ST} 
    \end{subfigure}
    \begin{subfigure}{0.137\linewidth}
        \includegraphics[width=\linewidth]{img2/final_train_1272_LFINet.png}
        \caption{LFINet}
    \end{subfigure}

    \caption{Visual comparison of the ablation study.}
    \label{fig:ablation}
\end{figure*}

\section{Results}

\subsection{Datasets}\label{AA}
\begin{figure}[t]
    \centering
    \includegraphics[width=\columnwidth]{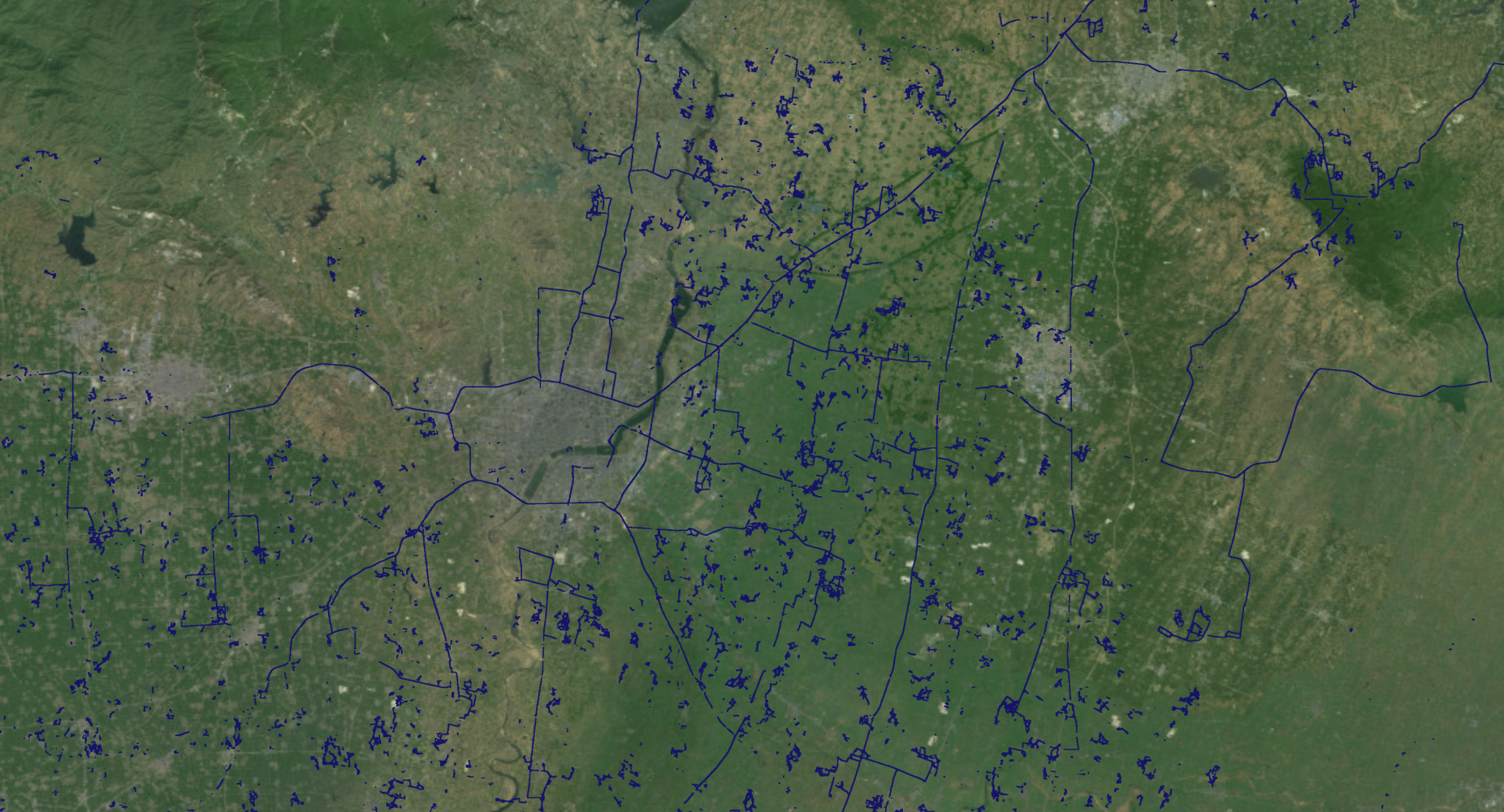}
    \caption{Road network results in Nanyang, Henan, China.}
    \label{fig:nanyang}
\end{figure}

To ensure the reliability and practical applicability of our method, we utilized a comprehensive dataset of real-world agricultural machinery trajectories. The primary data were collected during the peak wheat harvesting season in June 2021 across Henan Province. These trajectories record high-precision global navigation satellite system (GNSS) coordinates, speed, and heading information generated during actual field operations, capturing complex realistic scenarios such as turning maneuvers and road transits.
After converting the raw GNSS trajectories into image data, we employed data augmentation (e.g., rotation and flipping) to expand sample diversity. The augmented dataset was then stratified into a training set of 4,792 samples and a test set of 1,272 samples. Furthermore, we collected an additional 400 trajectory samples from Nanyang City, Henan Province, in June 2023 as an independent test set. The results of the road network construction, taking Nanyang, China, as a case study, are illustrated in Fig. \ref{fig:nanyang}.

\subsection{Experimental Setup}
LFINet was trained on an RTX 4090 GPU for 300 epochs using the Adam optimizer (learning rate $10^{-4}$, batch size 16). To handle class imbalance, we used a joint loss: $\mathcal{L} = \mathcal{L}_{\text{Dice}} + \mathcal{L}_{\text{BCE}}$. Performance was evaluated using Precision (Prec.), Recall, F1-score, IoU, and PSNR.

\subsection{Quantitative evaluation on different models}
To systematically evaluate our method, we compared LFINet with ten state-of-the-art models. These baselines can be grouped into three categories: (1) CNN-based semantic segmentation models (AD-HRNet~\cite{b3}, NL-LinkNet~\cite{b4}, Spd-Linknet~\cite{b5}, DeepMG~\cite{b1}, MADSNet~\cite{b9}); (2) Generative adversarial networks for image translation (T2R2M~\cite{b2}, T2R-pix2pix~\cite{b6}, T2R-GAN~\cite{b7}); and (3) Transformer and Mamba-based architectures (SAM-Road~\cite{b8}, SCSegamba~\cite{b10}). 

As presented in Table \ref{tab:comparison}, LFINet achieves state-of-the-art performance across most evaluation metrics, notably reaching a Precision of 98.60\% and an IoU of 86.12\%.

Compared to CNN-based models, LFINet shows a substantial lead in IoU (exceeding them by approximately 6--9\%). CNN-based backbones rely on continuous downsampling, which acts as a low-pass filter that irreversibly blurs the high-frequency local topological details of slender roads. By contrast, our LMS explicitly decouples these components at the input stage, ensuring that fine-grained geometric priors are preserved. While generative models demonstrate competitive PSNR due to their image-to-image translation nature, LFINet achieves a higher F1-score (92.54\%). This indicates that our ``decompose-interact-reconstruct'' strategy is less prone to generating ``hallucinated'' road segments in noisy areas than pure GAN-based synthesis. 

To highlight practical feasibility, we also evaluated the computational complexity of the models (\ref{tab:comparison}). LFINet maintains a  parameter count (22.11M), lower than many heavy models like SAM-Road (86.60M) and DeepMG (63.08M). Although its FLOPs (65.64G) are higher than lightweight CNNs, LFINet offers a superior trade-off between computational cost and topological accuracy.

\subsection{Qualitative Comparison}
To intuitively evaluate the perceptual quality and topological integrity of the extracted road networks, we present visual comparisons of a representative agricultural scenario in Fig. \ref{fig:seven}. 
As highlighted by the red bounding boxes, existing methods struggle to handle the structural discontinuity caused by sparse sampling. In the upper-left region, baseline models fail to bridge the gap between discrete trajectory points, resulting in fragmented segments. Similarly, in the central intersection area, baseline models produce blurred boundaries and miss critical connectivity links due to interference from field operation noise. This indicates that standard convolutions lack the capability to capture long-range dependencies required for repairing topological fractures.
In contrast, our method outperforms existing approaches, delivering more accurate, coherent, and robust road network
reconstructions.

\subsection{Ablation study}
To validate the contribution of each component, we conducted a comprehensive ablation study (Table \ref{tab:ablation}).
Replacing the LMS with convolutional layers results in a performance drop across all metrics, with IoU decreasing by 1.88\%. 
While removing the HFB leads to a decline in F1-score (91.90\%) and PSNR (30.38 dB),the absence of the Spatial Transformer (w/o ST) significantly degrades performance (IoU 83.28\%, PSNR 29.82 dB), underscoring the criticality of global semantic modeling.
Without this capability for capturing long-range dependencies, the model fails to connect fragmented segments.
When the FGM is removed, the IoU drops sharply by 3.50\%, and PSNR falls from 31.14 dB to 28.92 dB. In the absence of FGM, high-frequency details lack the ``filter'' provided by low-frequency semantic guidance, leading to the accumulation of background noise and a loss of reconstruction fidelity.
The ``w/o PRD'' variant (replaced by simple convolutions) shows the most drastic collapse, with IoU plummeting to 66.92\%. 
Visual comparisons (Fig. \ref{fig:ablation}) further corroborate that all ablated variants exhibit varying degrees of topological fractures, particularly in regions with sparse trajectory coverage.

\section{Conclusion}

To extract rural thematic road networks from agricultural trajectories, we proposed LFINet, a frequency-aware framework. LFINet achieves this by explicitly decoupling high-frequency structural components from low-frequency semantic representations via Laplacian pyramid decomposition and employing cross-frequency interaction. Experimental results demonstrate that LFINet achieves robust and accurate road extraction, establishing a new state-of-the-art performance compared to existing baselines. Future research will address several limitations. First, we will evaluate the model's generalization across diverse topographies (e.g., hilly terrains) beyond our current flat-terrain datasets. And statistical variance analyses will be conducted to validate robustness. Finally, graph-based vectorization will be explored to enhance topological completeness.

\end{document}